# Fuzzy Controller Design for Assisted Omni-Directional Treadmill Therapy

\* Atif Ali Khan, Oumair Naseer, Daciana Iliescu, Evor Hines,
School of Engineering, University of Warwick, Coventry, CV4 7AL, United K
Atif.Khan@warwick.ac.uk, O.naseer@warwick.ac.uk
D.D.Iliescu@warwick.ac.uk, E.L.Hines@warwick.ac.uk

*Abstract—* **One of the defining characteristic of human being is their ability to walk upright. Loss or restriction of such ability whether due to the accident, spine problem, stroke or other neurological injuries can cause tremendous stress on the patients and hence will contribute negatively to their quality of life. Modern research shows that physical exercise is very important for maintaining physical fitness and adopting a healthier life style. In modern days treadmill is widely used for physical exercises and training which enables the user to set up an exercise regime that can be adhered to irrespective of the weather conditions. Among the users of treadmills today are medical facilities such as hospitals, rehabilitation centers, medical and physiotherapy clinics etc. The process of assisted training or doing rehabilitation exercise through treadmill is referred to as treadmill therapy. A modern treadmill is an automated machine having some built in functions and predefined features. Most of the treadmills used today are one dimensional and user can only walk in one direction. This paper presents the idea of using omnidirectional treadmills which will be more appealing to the patients as they can walk in any direction, hence encouraging them to do exercises more frequently. This paper proposes a fuzzy control design and possible implementation strategy to assist patients in treadmill therapy. By intelligently controlling the safety belt attached to the treadmill user, one can help them steering left, right or in any direction. The use of intelligent treadmill therapy can help patients to improve their walking ability without being continuously supervised by the specialists. The patients can walk freely within a limited space and the support system will provide continuous evaluation of their position and can adjust the control parameters of treadmill accordingly to provide best possible assistance.**

*Keywords- Omni-directional treadmill, fuzzy logic, back pain, intelligent control, treadmill therpy.*

## I. INTRODUCTION

Treadmills are the most popular home exercise equipment. The treadmill ranked number one among exercise machines in burning calories (up to 865 an hour) at a perceived exertion intensity levels according to a 1996 study at the Medical College of Wisconsin and the Zablocki VA Medical Centre in Milwaukee. Treadmill is the most used of all indoor aerobic fitness equipment. Walking is one of the best forms of exercise and owning a treadmill makes sticking with your walking routine easier. If time, personal safety, allergies, or weather conditions, limit your outdoor walking a treadmill can be a great asset. Applications of treadmill include Exercise, Training, Physical Rehabilitation, Post Traumatic Stress Disorder Treatment, Motion Capture, Fully Immersive Gaming, Virtual Reality Simulations and Dog Training etc. It can be helpful in adopting a healthier lifestyle for back pain patients and the people who are less able to walk freely.

Traditional physical therapy is beneficial in restoring mobility in individuals who have sustained spinal injury or have less ability to walk but residual limitations often persist. Robotic technologies may offer opportunities for further gains. Walking recovery programs that use body-weight-supported treadmills are used in rehabilitation centers around the world offering patients a means of improving functional walking [1]. In a treadmill theraphy, patient walks on a treadmill under the supervison of a specialist with safty belts attached in case patient falls over. Medical specialist (often called physical therapist or physiotherapist) contineously monitors the patients moment and helps them during the treadmil walking session. A demonstration of conventioal treadmill theaphy is shown un the figure 1 below.

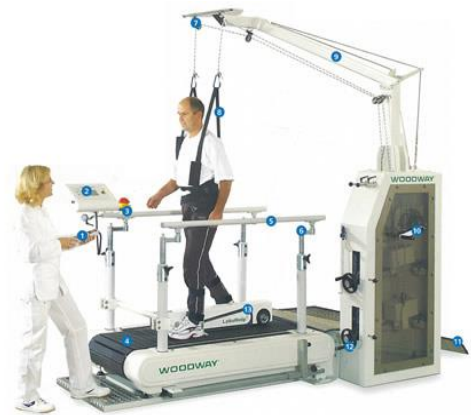

Figure 1, Demostration of assisted treadmill therphy
(Cortsey: Locomotion Therapy Product by Bio-Med Inc.)

## II. OMNIDIRECTIONAL TREADMILLS

One of the most popular types of home exercise equipment is the treadmill, which provides a straightforward, efficient aerobic workout. Treadmills are usually used for cardio workouts, walking in a one way or linear direction. An omnidirectional treadmill, or ODT, is a device that allows a person to perform locomotive motion in any direction. The ability to move in any direction is how these treadmills differ from their basic counterparts that permit only unidirectional locomotion. Virtual space devices are working on a concept that would make treadmills omni-directional, making the treadmill sense which way you are moving and adapting to that position [2]. The Omni-Directional Treadmill (ODT) is a revolutionary device for locomotion in large-scale virtual environments. Figure 2 below shows a person walking on an omnidirectional treadmill (virtual reality) with 3D glasses.

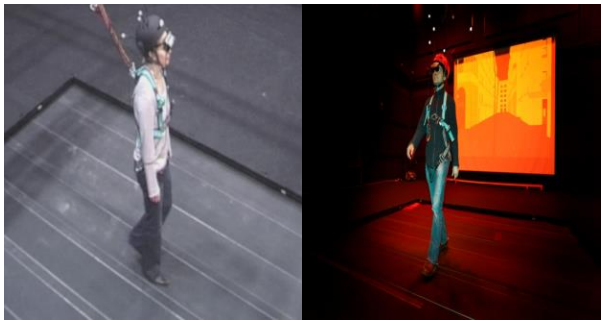

Figure 2, Real time Omni directional treadmill walking

Omnidirectional treadmills are employed in immersive virtual environment implementations to allow unencumbered movement within the virtual space through user self-motion. ODT's first appeared in 1997. There have been several designs aiming at providing the person with the natural feeling of walk [3]. Figure 3 and 4 below, shows different type of omnidirectional treadmills used for different kind of applications.

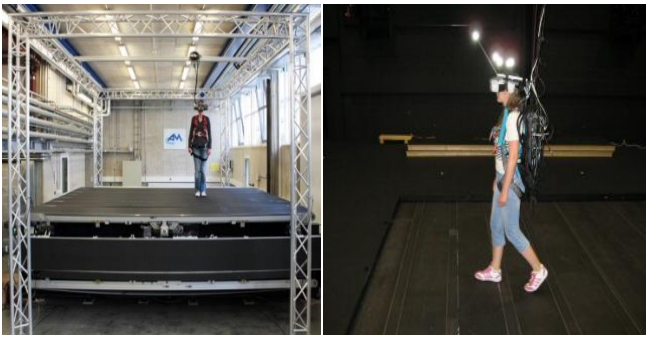

Figure 3, The Cyberwalk omni-directional treadmill at the Max Planck Institute for Biological Cybernetics

Omnidirectional treadmills were first professionally used by US army for the training purposes. This unit employed a belt made from plastic, toothed rollers. It employed mechanical tracking using a body harness, and included provisions for pneumatically powered, full-body force feedback for haptic and force display [4].

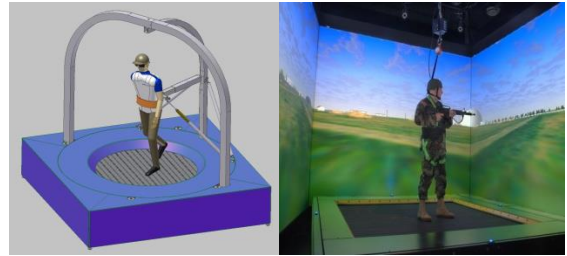

Figure 4, US Army Research Lab's ODT with CAVE Graphics

## III. TARGETED THERAPY FOR TARGETED PATIENTS

One of the growing problems now day is back pain which restricts the mobility of a person. Previously it was believed that a complete bed rest helps patient to overcome back pain. Modern research however shows that instead of a complete bed rest, physical exercise is more helpful to overcome back pain. A study carried out in 2002 [22] shows that treadmill therapy is helpful for pain relief and functional improvement of back pain patients.

### A. Back Pain

Back pain is usually associated with the Spine disorder. It is the second most common reason for visits to the doctor's clinic, outnumbered only by the upper-respiratory infections. [5] Back pain is one of the most common reasons for missed work. One-half of all working Americans admit to having back pain symptoms each year [5]. Back pain is very common; according to a survey published in 2000. Almost half the adult population of the UK (49%) report low back pain lasting for at least 24 hours at some time in the year [6]. In industrialized countries, it is estimated that four out of every five adults (80%) will experience back pain at some stage in their life [6]. The number of people with back pain increases with advancing age, starting in school children and peaking in adults of 35 to 55 years of age. Back pain is just as common in adolescents as in adults [7].

In most cases it is very difficult to identify a single cause for back pain. In about 85% of back pain sufferers no clear pathology can be identified [8]. The following factors could contribute to back pain; having had back pain in the past, smoking and obesity [7], physical factors such as heavy physical work, frequent bending, twisting, lifting, pulling, pushing, repetitive work, static posture and vibrations [10]. Psychosocial factors such as stress, anxiety, depression, job satisfaction, mental stress [7, 11]. The National Health Service (NHS) UK spends more than £1 billion per year on back pain related costs, this includes: £512 million on hospital costs for back pain patients, £141 million on GP consultations for back pain, £150.6 million on physiotherapy treatments for back pain [9]. In the private healthcare sector £565 million is spent on back pain every year. This brings the healthcare costs for back pain to a total of £1.6 billion per year [9]. In addition there are other (indirect) costs. The Health and Safety Executive estimates that musculoskeletal disorders, which include back pain cost UK employers between £590 million and £624 million per year [12]. The total cost of back pain corresponds to between 1% and 2% of gross national product (GDP) [13]. The charity BackCare estimates that back pain costs the NHS, business and the economy over £5 billion a year. Other European countries report similar high costs; back pain related costs in The Netherlands in 1991 were more than 4 billion euro. For Sweden in 1995 these were more than 2 billion euro [14]. Americans spend at least $50 billion each year on back pain and that's just for the more easily identified costs [15]. Nearly 5 million working days were lost as a result of back pain in 2003-04. Each person suffering from such a condition took an estimated 17.4 days off work on average in this period [16]. Back pain is the number two reason for long term sickness in much of the UK. In manual labor jobs, back pain is the number one reason [17].

*B. Treadmill Theraphy*

Treadmill walking training program help patients who have difficulty in upright mobility to improve their walking ability during the last two decades. Studying and understanding the phenomenon of treadmill walking could be useful for the rehabilitation applicant. Robotic assisted treadmill therapy is a safe method to enable longer periods of gait therapy in children and adolescents with gait disorders [18]. A research carried out in 2011 showed that persons with chronic, motor-incomplete SCI can improve walking ability and psychological well-being following a concentrated period of ambulation therapy, regardless of training method. Improvement in walking speed was associated with improved balance and muscle strength. In spite of the fact that we withheld any formal input of a physical therapist or gait expert from subjects in the device-specific training groups, these subjects did just as well as subjects receiving comprehensive PT for improving walking speed and strength. [19]

A comfortable, supportive harness is quick and easy to put on and off. Worn as a vest around the upper torso, the "Support Harness" provides patients with security and safety from falling. Versatility includes double shoulder point connections to spreader bar or, a single connection point for low ceiling applications with tall patients (6'3" on a treadmill or 6'11" on the floor). A pelvic support strap is included to keep the harness in place when limited body weight support is intended [20]. An example of support harness system is shown in figure 5 below.

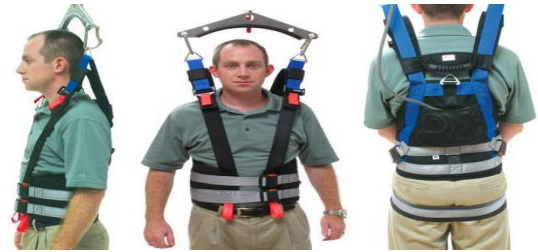

Figure 5, Biodex Support Harness DLX,
BIODEX Medical Systems, Inc. New York.

The aim of this study is to design an efficient and intelligent control system for the harness which can automatically adjust itself to provide maximum safety and assistant to the patient during the treadmill walking session.

Treadmills therapy is usually recommended for Stroke, Spinal Cord, Head Injury, Amputees, Orthopedic, Neurologic, Vestibular and Older Adult Patients. In 1995, a study was conducted to study and clarify the extent to which elderly people have difficulties in mobility, and determine their association with socioeconomic factors, dwelling environment and use of services. The study groups were composed of two random samples of 800 persons aged 65-74 and 75-84, respectively. The respondents were asked to assess their ability to get about the house, negotiate stairs and walk outdoors, as well as manage certain physical exercise tasks. Difficulties in getting about outdoors were found most frequently among the women in the older age group (52% reported difficulties), and least frequently among the women in the younger age group (23%). Logistic regression analyses showed that difficulties in getting about outdoors were significantly explained by length of education and defects in the dwelling environment. It was concluded in that research that the difficulties in mobility among elderly people, especially among elderly women, should be reduced more actively either by improving their physical abilities or by developing compensation strategies for their own use or in regard to the environment [21]. So treadmill walking can be a good alternative to cope with the weather conditions which hinders outdoor mobility in elderly peoples. A study carried out in 2002 shows that treadmill therapy is helpful for pain relief and functional improvement of back pain patients [22]. An assisted treadmill walking is shown in figure 6 below.

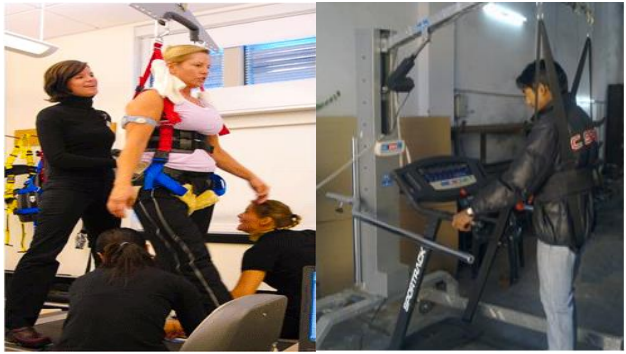

Figure 6, Left: Students in Physical Therapy Program, University of North Carolina using supported treadmill therapy to regain a patient's ability to walk. Right: Locomotion Therapy Product by Bio-Med Inc.

### IV. METHODOLGY

A basic system diagram is shown in figure 7 below. Fuzzy logic is used for the control strategy. Fuzzy logic is widely used in machine control. The term itself inspires certain scepticism, sounding equivalent to "half-baked logic" or "bogus logic", but the "fuzzy" part does not refer to a lack of rigour in the method, rather to the fact that the logic involved can deal with concepts that cannot be expressed as "true" or "false" but rather as "partially true". Although genetic algorithms and neural networks can perform just as well as fuzzy logic in many cases, fuzzy logic has the advantage that the solution to the problem can be cast in terms that human operators can understand, so that their experience can be used in the design of the controller. This makes it easier to mechanize tasks that are already successfully performed by humans. Fuzzy logic controller gets the information against the position of user on the treadmill and computes the best parameter setting for treadmill support (belt) to keep user on track. The decision of parameter setting is computed on the basis of predefine fuzzy set of rules. The model is recursive that is loop. Controller checks the position information after certain amount of time and decides the appropriate action on the basis of current position. The goal of the controller is to assist user in walking and make sure that user never crosses the critical limit in any direction.

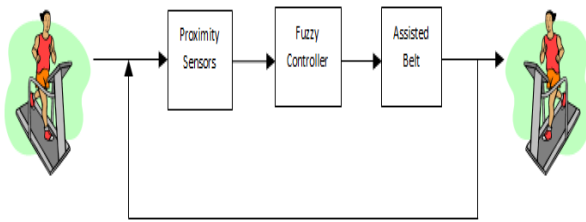

Figure 7 Overview of the system

### V. FUZZY CONTROLLER

A fuzzy control system is based on fuzzy logic. A mathematical system that analyses analog input values in terms of logical variables that take on continuous values between 0 and 1, in contrast to classical or digital logic, which operates on discrete values of either 1 or 0 (true or false, respectively). Classic control theory uses a mathematical model to define a relationship that transforms the desired state (requested) and observed state (measured) of the system into an input or inputs that will alter the future state of that system [23]. The purpose of control is to influence the behaviour of a system by changing an input or inputs to that system according to a rule or set of rules that model how the system operates.

In the proposed mode, fuzzy control design is presented for controlling the safety belt attached to the omnidirectional treadmill user. By efficiently controlling the safety belt, assistance can be provided to the patients during the walking session on the omni-treadmill. If the person goes too far in either direction, there is a chance for over running the treadmill and can eventually cause physical damage. To overcome such situation, fuzzy controller monitors the patient's location continuously and if patient goes too far in any direction and reaches the critical limit, the controller drags the safety belt in opposite direction to avoid overrunning the track. The overall goal of the controller is to keep patient in or near the centre of the treadmill which is the safest place. The input to fuzzy controller is the patient's actual position on the treadmill and the output is the steering of the assisted safety belt attached to the patient. A fuzzy controller design is shown in figure 8 below.

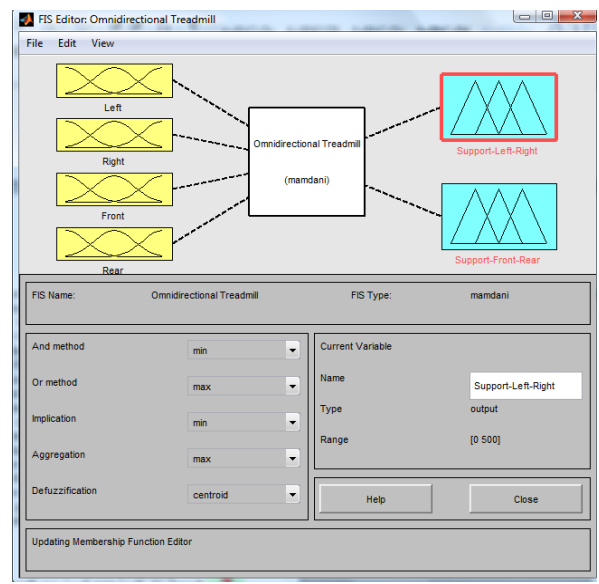

Figure 8, Fuzzy controller input and outputs

The membership functions of the inputs are categorised into two groups as "far" and "near". For the simplicity, only two membership functions are used. The number of membership functions can be increased to get more fine results but on the other hand it will make system more complex and computationally less efficient. The range of inputs are set to 0 -500. The output is the steering direction (the direction in which assistance is required). The range of output is set from 0-500. Input and output membership functions are shown in figure 9 below.

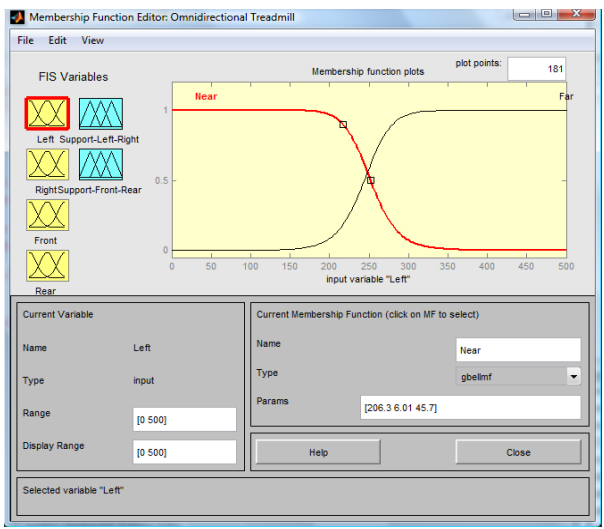

Figure 9, Input and output membership functions

The fuzzy controller operates on the principle of if-then rules called fuzzy set of rules. For the proposed model, 10 rules were used which are given below.

If **front** is **far** and **rear** is **near** then **support** is **front**.

If **front** is **near** and **rear** is **far** then **support** is **rear**.

If **left** is **near** and **front** is **near** then **support** is **right** and **rear**.

If **left** is **near** and **rear** is **near** then **support** is **right** and **front**.

If **right** is **near** and **front** is **near** then **support** is **left** and **rear**.

If **right** is **near** and **rear** is **near** then **support** is **left** and **front**.

If **left** is **far** and **front** is **far** then **support** is **left** and **front**.

If **left** is **far** and **rear** is **far** then **support** is **left** and **rear**.

If **right** is **far** and **front** is **far** then **support** is **right** and **front**.

If **right** is **far** and **rear** is **far** then **support** is **right** and **rear**.

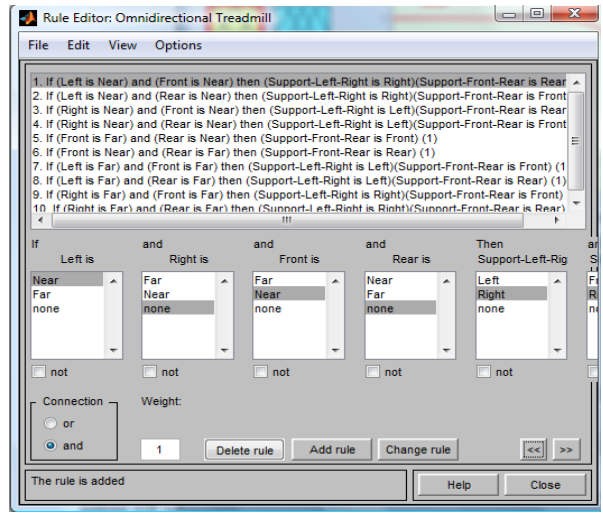

Figure 10, Fuzzy rule editor

The fuzzy rule base is shown in figure 11 below. Fuzzy rule base provides the probability or the firing strength of each rule against the input and output variables.

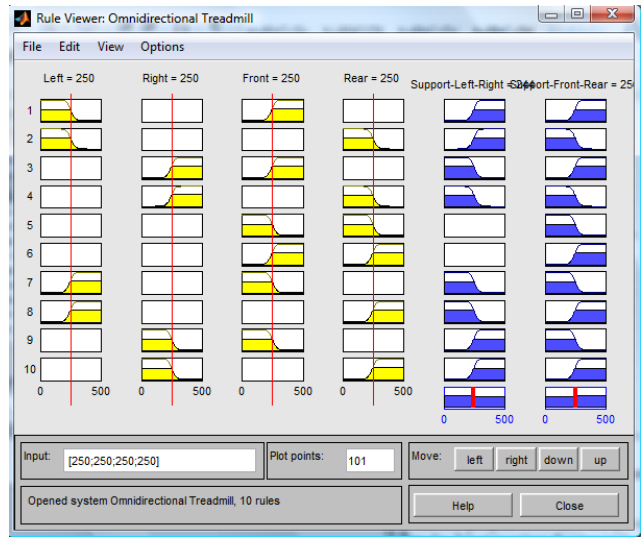

Figure 11, Fuzzy rule base

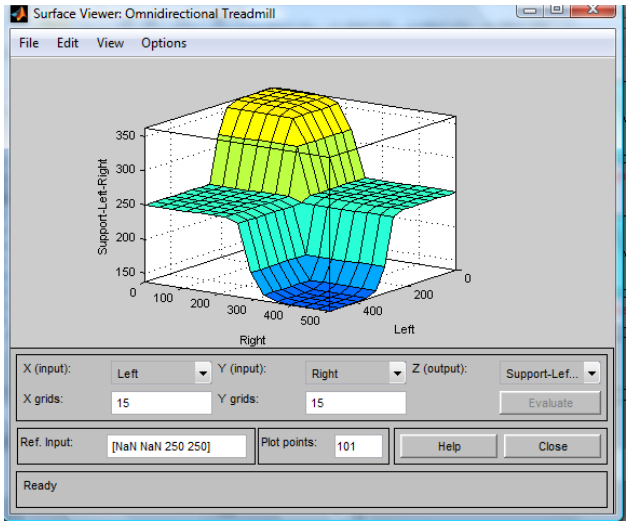

Figure 12, Surface view of the overall design

Surface view provides the visual description of the relationship between the input and output variables.

## VI. SIMULATION RESULTS

Some dummy tracks were created in matlab using x-y coordinates. A 500x500 two dimensional (x-y coordinates) track was created by varying the values of horizontal and vertical axis. The efficiency of controller was tested by running simulation and traversing patients through the dummy tracks. Matlab and Simulink were used for the controller design and simulations. The blue dots represent the original path followed by the patient. Figure 13 show that patient moves out of the limit of the track and the program has crashed. The conventional support system attached to patient will just lift the patient in such case to avoid any injury.

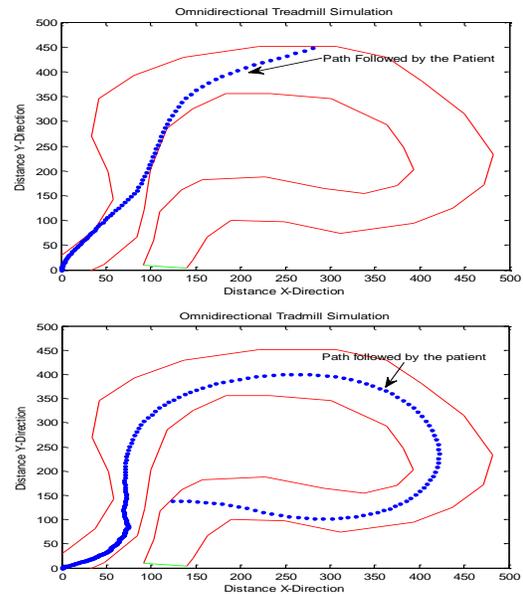

Figure 13, Path Followed by the patient without supervision

Whereas in figure 14 below; it can be seen that in the presence of fuzzy logic controller, the patient is able to move through the full length of track. Fuzzy logic controlled support provides assistance if the patient comes too close to the boundaries. Comparing figures 13 above and 14 below, it can be seen that fuzzy logic controller is able to avoid patient from going off the track. In fig 14, when patient reached the critical limit and tends to go off track, fuzzy logic controller is good enough to keep patient on track by providing assisted in steering towards right side.

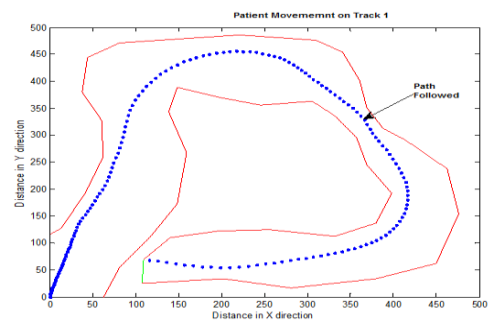

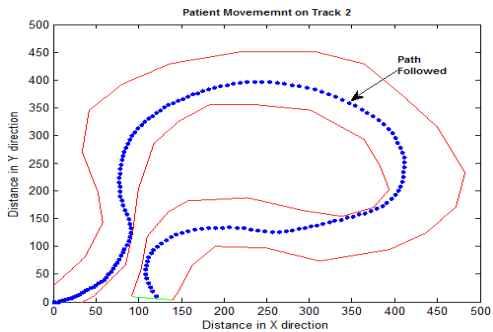

Figure 14, Path followed by the patient with fuzzy controller supervision

In this proposed model, the controller program does not have any pre-defined information of tracks. It just monitors the current location of the patient with the treadmill boundaries in all four directions. If patient reaches any critical limit on left, right, front or rear, the controller assist patient to avoid going out the treadmill track.

## VII. CONCLUSION

The treadmill is a relatively an easy piece of exercise equipment to use. A model based on fuzzy logic for controlling the walking movement has been proposed in this paper. The aims and objectives were achieved successfully. The intelligent control strategy and intuitive support method is applied in order to control patient walk in a natural manner. Fuzzy logic based expert system is developed to choose suitable control strategy for different state during walking. In conventional system, when dealing with patients with walking difficulties, a physiotherapist is required to monitor the patient using treadmill consistently. In this proposed system, overall work load of a physiotherapist can be minimized to a great extent. An intelligence expert system serves the basic role of a supervisor and consistently monitors the patient moment and adjusts the different parameters of the treadmill support accordingly to give a maximum confidence and support to the treadmill user. The fuzzy controller can be synchronised with the virtual tracks to monitor the efficiency of the patient against certain track and can help them traversing through the specified track more efficiently and safely. By adapting this intelligent control strategy the fear factor of falling over for the patients can be eliminated or minimised at least. This will encourage them to do exercise frequently and more willingly.

## VIII. FUTURE RECOMMENDATIONS

Walking is known as one of the most universal and yet most complex for of all human activities. The thoroughly understanding of the walk cycle is important to design and implement an appropriate control model. The joints, hip, knee, wrist, neck and ankle (the junctions linking the lower limbs and upper limbs which provide energy to drive the segments activities) can be used to design and develop a modern and more sophisticated expert system.